\PassOptionsToPackage{dvipsnames}{xcolor}

\documentclass[10pt,journal,compsoc]{IEEEtran}
\usepackage{cite}
\usepackage{graphicx}
\usepackage{tikz}
\usepackage{comment}
\usepackage{amsmath,amssymb}
\usepackage{hyperref}
\usepackage{color}
\usepackage[accsupp]{axessibility}
\usepackage{multirow}
\usepackage{makecell}
\usepackage{times}
\usepackage{epsfig}
\usepackage{amsfonts}
\usepackage{bbm}
\usepackage{mathabx}
\usepackage{array}
\usepackage{pifont}
\usepackage{scalerel,xparse}
\usepackage{booktabs}
\usepackage{tikz}
\usetikzlibrary{backgrounds,mindmap}
\usetikzlibrary{calc,positioning,intersections}
\usepackage{graphicx}
\usepackage{mathtools}
\NewDocumentCommand\emojifire{}{\scalerel*{\includegraphics{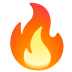}}{X}}
\NewDocumentCommand\emojisnow{}{\scalerel*{\includegraphics{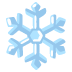}}{X}}
\NewDocumentCommand\emojistar{}{\scalerel*{\includegraphics{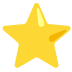}}{X}}
\NewDocumentCommand\emojitarget{}{\scalerel*{\includegraphics{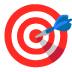}}{X}}
\NewDocumentCommand\emojihand{}{\scalerel*{\includegraphics{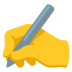}}{X}}
\NewDocumentCommand\emojigpt{}{\scalerel*{\includegraphics{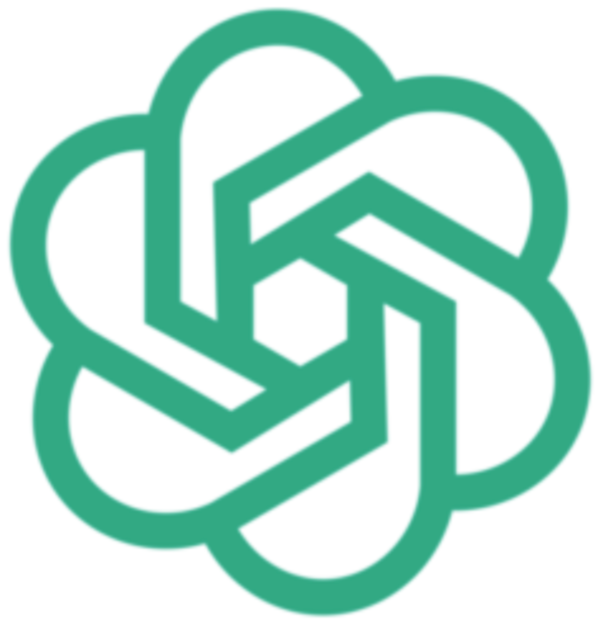}}{X}}
\NewDocumentCommand\emojiauto{}{\scalerel*{\includegraphics{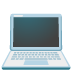}}{X}}

\newcommand{\cmark}{\ding{51}}%
\newcommand{\xmark}{\ding{55}}%
\usepackage{tabu}
\newcolumntype{M}[1]{>{\centering\arraybackslash}m{#1}}
\newcolumntype{L}[1]{>{\flushleft\arraybackslash}m{#1}}
\definecolor{mygray}{gray}{0.70}

\ifCLASSINFOpdf
\else
\fi
\begin{document}

\newcommand{\etal}{\textit{et al. }}
\newcommand{\eg}{\textit{e.g.}}
\newcommand{\ie}{\textit{i.e.}}

\title{Generative Physical AI in Vision: A Survey}

\author{
    Daochang~Liu, 
    Junyu~Zhang,
    Anh-Dung~Dinh,
    Eunbyung~Park,
    Shichao~Zhang,
    Ajmal~Mian,
    Mubarak~Shah,
    Chang~Xu
\IEEEcompsocitemizethanks{

\IEEEcompsocthanksitem D. Liu and A. Mian are with School of Physics, Mathematics and Computing, The University of Western Australia, Crawley, WA 6009, Australia.
Email: daochang.liu@uwa.edu.au, ajmal.mian@uwa.edu.au

\IEEEcompsocthanksitem J. Zhang is with the Department of Electrical and Computer Engineering, Sungkyunkwan University, Suwon 16419, Korea.
E-mail: zhangjunyu@skku.edu

\IEEEcompsocthanksitem A. Dinh, and C. Xu are with School of Computer Science, Faculty of Engineering, The University of Sydney, Darlington, NSW 2008, Australia.
Email: adin6536@uni.sydney.edu.au, c.xu@sydney.edu.au

\IEEEcompsocthanksitem  E. Park is with the Department of Artificial Intelligence, Yonsei University, Seoul 03722, Korea.
E-mail: epark@yonsei.ac.kr

\IEEEcompsocthanksitem  S. Zhang is with the Guangxi Key Lab of Multi-Source Information Mining \& Security, Guangxi Normal University, Guilin 541004, China.
E-mail: zhangsc@csu.edu.cn

\IEEEcompsocthanksitem M. Shah is with the Center for Research in Computer Vision, University of Central Florida, Orlando, FL 32816-2365, USA.
E-mail: shah@crcv.ucf.edu


}
}

%
%

\markboth{Journal of \LaTeX\ Class Files,~Vol.~14, No.~8, August~2015}%
{Shell \MakeLowercase{\textit{et al.}}: Bare Demo of IEEEtran.cls for Computer Society Journals}
%



\IEEEtitleabstractindextext{%
\begin{abstract}
Generative Artificial Intelligence (AI) has rapidly advanced the field of computer vision by enabling machines to create and interpret visual data with unprecedented sophistication. 
This transformation builds upon a foundation of generative models to produce realistic images, videos, and 3D/4D content. 
Conventional generative models primarily focus on visual fidelity while often neglecting the physical plausibility of the generated content. 
This gap limits their effectiveness in applications that require adherence to real-world physical laws, such as robotics, autonomous systems, and scientific simulations. 
As generative models evolve to increasingly integrate physical realism and dynamic simulation, their potential to function as ``world simulators" expands.
Therefore, the field of physics-aware generation in computer vision is rapidly growing, calling for a comprehensive survey to provide a structured analysis of current efforts.
To serve this purpose, the survey presents a systematic review, categorizing methods based on how they incorporate physical knowledge—either through explicit simulation or implicit learning. It also analyzes key paradigms, discusses evaluation protocols, and identifies future research directions. By offering a comprehensive overview, this survey aims to help future developments in physically grounded generation for computer vision.
The reviewed papers are summarized at \url{https://tinyurl.com/Physics-Aware-Generation}.
\end{abstract}

}

\maketitle

\IEEEdisplaynontitleabstractindextext

%
\IEEEpeerreviewmaketitle

\section{Introduction
\label{sec:introduction}}

\IEEEPARstart{G}{enerative} learning has long been a foundational pillar of computer vision, addressing key challenges in understanding, synthesizing, and manipulating visual data.
Over the past decade, this field has witnessed rapid evolution of diverse generative models, including Variational Autoencoders (VAEs)\cite{gu2022vector,Xing2024}, Generative Adversarial Networks (GANs)\cite{GAN}, Diffusion Models (DMs)\cite{DDIM,BeatsGAN,DDPM}, Neural Radiance Fields (NeRFs)\cite{gao2022nerf}, Gaussian Splatting (GS)\cite{Fei2024}, and Visual Autoregressive Models (VARs)\cite{Tian2024}. 
These models have consistently pushed the boundaries of generative learning, leveraging increasingly powerful architectures to capture the underlying distributions of visual data. 
The goal is to enable machines to reason about the visual world in ways that mirror human creativity and understanding, by imagining novel instances of visual content under unseen scenarios.
Among these advancements, diffusion models have emerged particularly noteworthy for their ability to produce high-fidelity outputs. By iteratively refining random noise through learned denoising processes, diffusion models demonstrate exceptional robustness and versatility, making them a cornerstone of recent generative methodologies. 
The applications of generative models span diverse modalities of visual contents, including image generation with semantics understanding, video generation with dynamic temporal understanding, 3D content generation with enhanced spatial understanding\cite{Cai2024b,Gao2024b,Xiang2024b}, and 4D contents with more complex spatiotemporal understanding\cite{Yuan2024,Wu2024a,Wimmer2024,Ren2023,Ling2024,Lee2024b,Pang2024}.
These advancements underscore the growing potential of generative learning in the broad landscape of increasingly sophisticated visual tasks.

\begin{figure}[t]
\begin{center}
\includegraphics[width=1\linewidth]{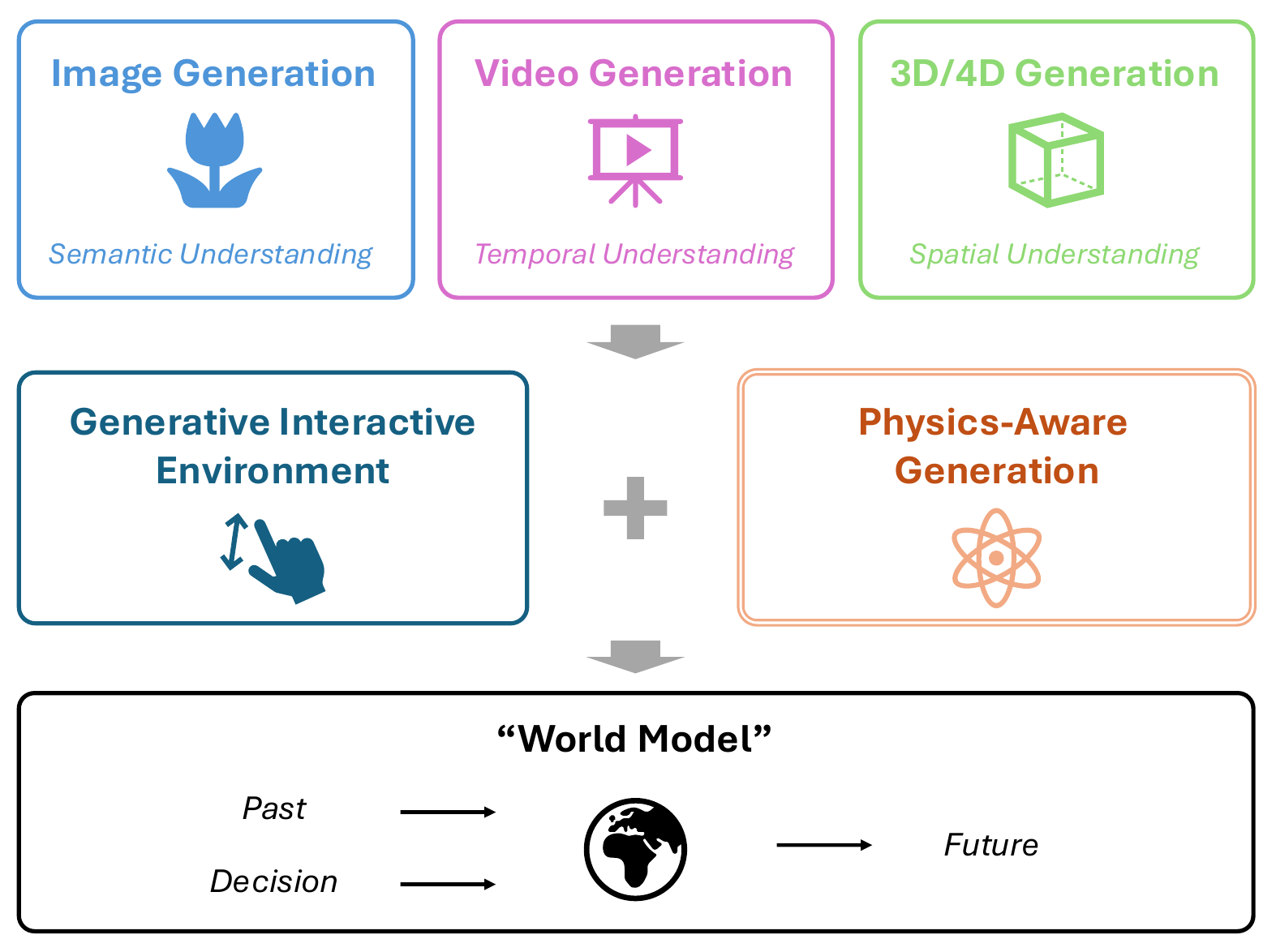}
\end{center}
   \caption{Overview of the progression of generative AI in vision and the role of physics-aware generation in building a world simulator. The figure shows the evolution from conventional focuses in generative tasks on semantic, temporal, and spatial understanding to incorporating interactivity and physical realism. Physics-aware generation enables models to simulate real-world dynamics, advancing toward general-purpose world models for various applications.}
\label{fig:introduction}
\end{figure}

\begin{figure*}[t]
\begin{center}
\includegraphics[width=1\linewidth]{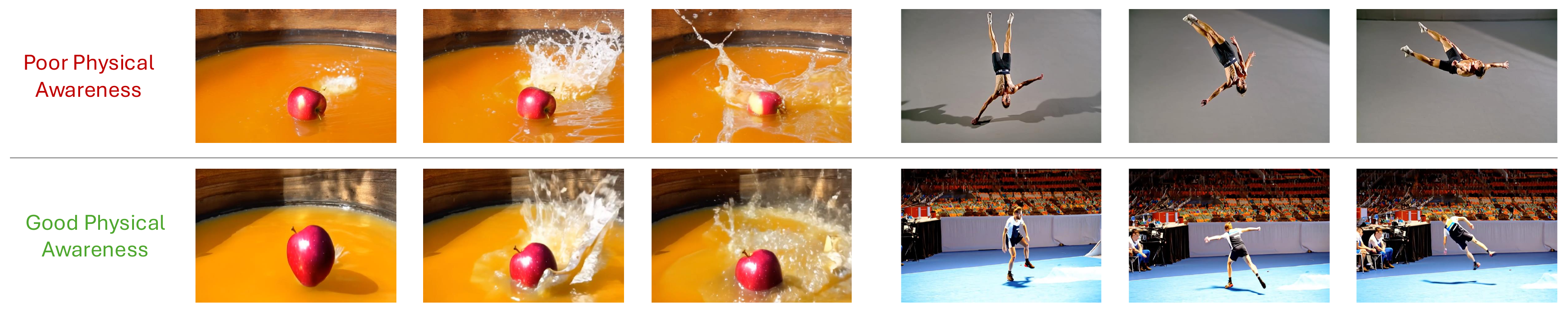}
\end{center}
   \caption{Examples of generated videos with poor and good physical awareness. Examples are from WISA~\cite{wang2025wisa} and Zhao et al~\cite{zhao2025synthetic}.}
\label{fig:intro-examples}
\end{figure*}

Among these visual modalities, video generation has recently gained significant attention in generative learning, presenting a far more challenging testbed for scaling up large generative models to handle higher dimensional data.
This complexity stems not only from the spatial intricacies within individual frames but also from the temporal coherence required across multiple frames.
Numerous video generation models have been developed and have captured widespread public interest, such as OpenAI's Sora\cite{Sora}, Google's Veo2\cite{Veo2}, Tencent's Hunyuan\cite{Hunyuan}, Kuaishou's Kling\cite{Kling}, and Nvidia's Cosmos~\cite{Cosmos}.
Video generation has been intensively studied in a range of formulations and settings, spanning from the vanilla unconditional generation\cite{PYoCo,Gupta2023}, to image-to-video generation\cite{Blattmann2023,Xing2023a,Yu2023b,Chen2023b,Kondratyuk2024,Guo2024a,Guo2024b,Hu2022}, text-to-video generation\cite{PYoCo,Blattmann2023,Wu2023a,Wang2023a,Gupta2023,Chen2023b,Girdhar2024,Singer2022,Kondratyuk2024}, video-to-video generation\cite{Wu2023b,Liu2024d}, and video editing or customization\cite{Wei2023,Ren2024b,Li2023c,Wang2023b}.
Each of these settings addresses unique challenges, from preserving temporal coherence to incorporating semantic guidance from textual or visual inputs.

\begin{table*}[t]
  \centering
  \begin{tabular}{||p{5cm}|p{1cm}|p{1cm}|p{1cm}|p{1cm}|p{1cm}|p{1cm}||}
 \hline
 & PS & PU & G & PUG & PAG-E & PAG-I  \\
 \hline
Inputs (Observations)           & \cmark & \cmark & $\circ$& $\circ$& $\circ$& $\circ$\\
Inputs (Physics)                & \cmark & $\circ$& $\circ$& \xmark & $\circ$& $\circ$\\
Outputs (Observations)          & \cmark & $\circ$& \cmark & \cmark & \cmark & \cmark \\
Outputs (Physics)               & $\circ$ & \cmark & $\circ$& \xmark & $\circ$& $\circ$\\
\hline
Explicit Physical Model         & \cmark & \cmark & $\circ$& \xmark & \cmark & \xmark\\
Understanding Physical World    & \cmark & \cmark & $\circ$& $\circ$ & \cmark & \cmark \\
\hline
  \end{tabular}
  \caption{Comparison of Concepts. \cmark: Yes. \xmark: No. $\circ$: Optional.\\
  PS: Physical simulation, PU: Physical understanding. G: Generation. PUG: Physics-unaware generation. PAG-E: Physics-aware generation with explicit physical simulation. PAG-I: Physics-aware generation without explicit physical simulation.}
  \label{tab:comparison_of_concepts}
\end{table*}

Video has a pivotal role in generative AI in computer vision, since the vast amount of video data available online encapsulates a rich repository of information about the real world, making video a medium through which generative AI can learn to model complex real-world phenomena.
In this context, video can be regarded as an implicit physical model of the world, offering the potential to bridge the digital and physical domains\cite{Yang2024b}, and paving the way for real-world decision-making.
Such a model could facilitate a large number of downstream tasks, including autonomous driving, scientific simulation, robotics\cite{Hafner2019,Yang2023b,Hafner2020,Hafner2023,Raad2024}, and other forms of embodied intelligence. 
To realize this potential, the generation process must be capable of interacting with external controls from humans or other systems.
This interactivity facilitates dynamic decision-making and the optimization of outcomes based on interactions, giving rise to what can be described as generative interactive environments\cite{Wang2024e,Yang2024b,Qin2024,Bruce2024}. 
Video generation has been incorporated with various interactive control signals, such as motion vectors or trajectories\cite{Blattmann2021b,Davtyan2024,Geng2024b,Chen2023d,Lu2024b}, hand masks\cite{Sudhakar2024}, latent actions\cite{Bruce2024,Menapace2021}, robot operations\cite{Yang2023b}, camera motions\cite{Menapace2022}, demonstrations\cite{Sun2024a}, and natural language descriptions\cite{Xiang2024a,Lai2023,Menapace2024}.
These interactive elements highlight the versatility and adaptability of generative video models, showing the potential to evolve into world models.

However, a critical gap remains in the transition from generation to robust world modeling, i.e., the ability to faithfully understand and replicate real-world physics\cite{AlTahan2024} (Fig.~\ref{fig:introduction}).
Current state-of-the-art models are primarily optimized for visual realism in the pixel space, rather than physical plausibility in the entity or concept space (Fig.~\ref{fig:intro-examples}).
For generative models to serve as physical world simulators, they must incorporate a deep understanding of physical laws, such as dynamics, causality, and material properties. 
This physics awareness is essential for going beyond generating visually appealing outputs to ensuring that the content also aligns with the constraints 
of the physical world.
The field of physics-aware generation is rapidly expanding, yet it remains fragmented, encompassing a wide range of diverse tasks, lacking clear definitions and standardized evaluation protocols.
Therefore, we provide this survey as a timely and comprehensive review of existing literature to streamline research efforts.
By examining existing efforts, we aim to highlight the progress made so far, provide a structure of paradigms, and identify potential future directions.

\textbf{Survey Scope.}
This is the first survey with a scope of generative models in computer vision that enhance the physical awareness of the generation {\em outputs}. 
We shift away from discussing 
methods that incorporate physical principles as prior knowledge or inductive biases in the model or neural architecture design, such as Physics-Informed Neural Networks (PINNs)\cite{Wang2024a,AwesomePINN}, even if the task is related to generative learning, e.g., \cite{Wu2024b,Su2023,Mao2024a}.
Moreover, we focus on generative tasks, and thus exclude image processing tasks such as de-blurring, de-hazing, and enhancement from our scope, though we notice that a large body of these works incorporate physics. 
Finally, to keep the survey fully focused on computer vision, pure graphics research involving physical simulation is also excluded. 







\textbf{Comparison to Other Surveys.}
This survey is orthogonal to existing surveys on physics-informed machine learning\cite{Karniadakis2021}, physics-informed computer vision\cite{Banerjee2023}, and physics-informed artificial intelligence\cite{Jiao2024}, since they emphasize the model design aspect with physical prior knowledge rather than enhancing the physical realism of generation outputs.
Our survey focuses on physics-aware generation, thus being distinct from existing surveys with general scopes on generative models\cite{LoaizaGanem2024}, diffusion models\cite{luo2022understanding,DiffusionSurvey}, video diffusion models\cite{Xing2023b}, diffusion-based video editing\cite{Sun2024b}.
Our survey also has a different scope compared to those in specific domains, such as human video or motion generation\cite{Lei2024,Xue2024b,Zhu2023}.
A recent, subsequent survey~\cite{lin2025exploring} has a similar scope, but from a different perspective of cognitive evolution processes of physics in video generation models.












\section{Formulation}

We first provide definitions of \textit{physics-aware generation} and related concepts, such as \textit{physical simulation} and \textit{physical understanding}. Based on these definitions, we further identify the common paradigms of incorporating physics into generative models in computer vision, to provide a structural perspective for later sections of this survey.

\subsection{Definitions}

Let $P_\theta$ denote a physical simulation model with physical parameters $\theta$, and $G$ represent a generative model, we can then provide the following definitions:

\textbf{Physics Simulation (PS)}: 

\begin{equation}
    P_\theta(X) \rightarrow X'.
\end{equation}

Physics simulation is the process evolving the input observation $X$ into the output observation $X'$ using the physical model $P_\theta$, where the observations $X,X'$ can be from different simulation time steps.

\textbf{Physics Understanding (PU)}: 

\begin{equation}
    X \rightarrow P_\theta.
\end{equation}

Physics understanding is the process inferring the underlying physical model $P_\theta$ from the observation $X$ such as video data. Physics understanding can also be conducted to only infer the physical parameters $\theta$ given a predefined physical model $P$.

\textbf{Generation (G)}: 

\begin{equation}
    G(X) \rightarrow X'.
\end{equation}

Generation is the process creating new content $X'$ from the input condition $X$ using the generative model $G$. The input and output can take various forms of modalities depending on the specific task.
If the generation process is general-purposed and does not necessarily involve a strong understanding of the physical world, we call it as physics-unaware generation (PUG).



\subsection{Formulation of Physics-Aware Generation}

Based on the above concepts, we further define the concept of physics-aware generation (PAG) as the generation process with a strong understanding of real-world physics.
We divide physics-aware generation into two major categories, i.e., physics-aware generation with explicit physical simulation (PAG-E), and physics-aware generation with implicit learning (PAG-I). 
This division is based on whether the generative model utilizes a physical simulation model explicitly to improve physical awareness.
Table~\ref{tab:comparison_of_concepts} provides a comparison of these concepts.

Within the category of physics-aware generation with explicit physical simulation (PAG-E), we further identify the following common paradigms of how the physical simulation is incorporated into the generative model.

\begin{figure}[h]
\begin{center}
\includegraphics[width=1\linewidth]{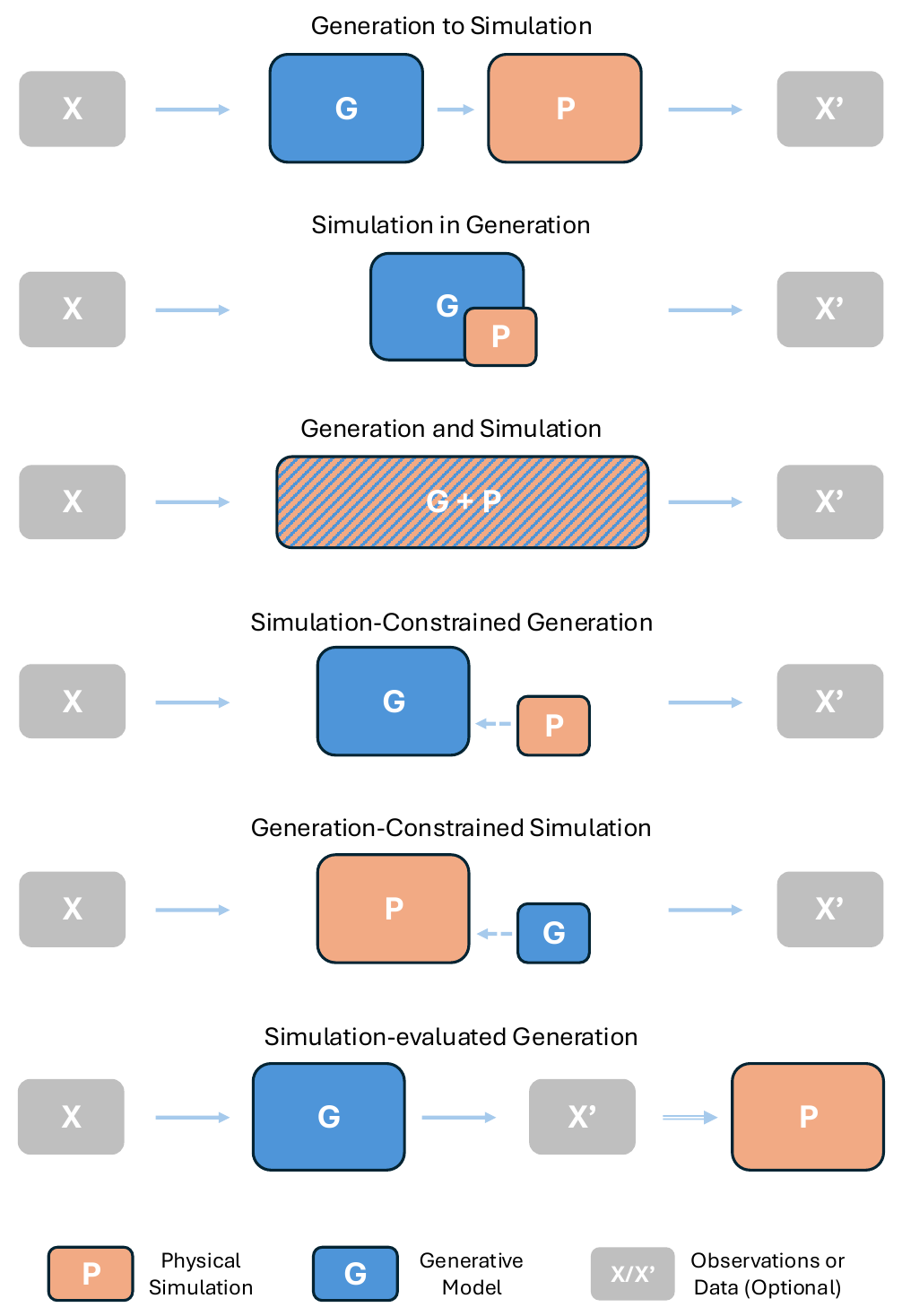}
\end{center}
   \caption{Illustration of six paradigms for integrating explicit physical simulation into generative models, highlighting structural differences in terms of the interaction between the generative process $G$ and the physics simulation model $P$.}
\label{fig:paradigms}
\end{figure}



\noindent 
\textbf{Generation to Simulation (GtS)}: This paradigm is a sequential composition where the simulation process follows the generation process:

\begin{equation}
P_\theta(G(X)) \rightarrow X'.
\end{equation}

\noindent 
\textbf{Simulation in Generation (SiG)}: The simulation model is incorporated as a part or a sub-module of the generation model:
\begin{equation}
G_{P_\theta}(X) \rightarrow X'.
\end{equation}

\noindent 
\textbf{Generation and Simulation (GnS)}: The generation and the simulation happen simultaneously, such as by using a shared model $M$:
\begin{equation}
M_{P_\theta, G}(X) \rightarrow X'.
\end{equation}


\noindent 
\textbf{Simulation-Constrained Generation (ScG)}: The simulation model is used to provide constraints or knowledge to the generation model:
\begin{equation}
G(X) \rightarrow X' \;\; \mathrm{subject \; to} \;\; P_\theta(X) \rightarrow X'.
\end{equation}

\noindent 
\textbf{Generation-Constrained Simulation (GcS)}: The generation model is used to provide constraints or knowledge to the simulation model:
\begin{equation}
P_\theta(X) \rightarrow X' \;\; \mathrm{subject \; to} \;\; G(X) \rightarrow X' .
\end{equation}

\noindent 
\textbf{Simulation-evaluated Generation (SeG)}: The physical simulation is used to evaluate the generative model or the generated content is intended for deployment in simulation environments.

These paradigms are illustrated in Fig.~\ref{fig:paradigms}. 
We cover physical simulation in Section~\ref{sec:simulation}, physical understanding in Section~\ref{sec:understanding}, physical-aware generation in Section~\ref{sec:generation}, and physical evaluation in Section~\ref{sec:evaluation}.


\begin{figure*}[h]
\begin{center}
\includegraphics[width=1\linewidth]{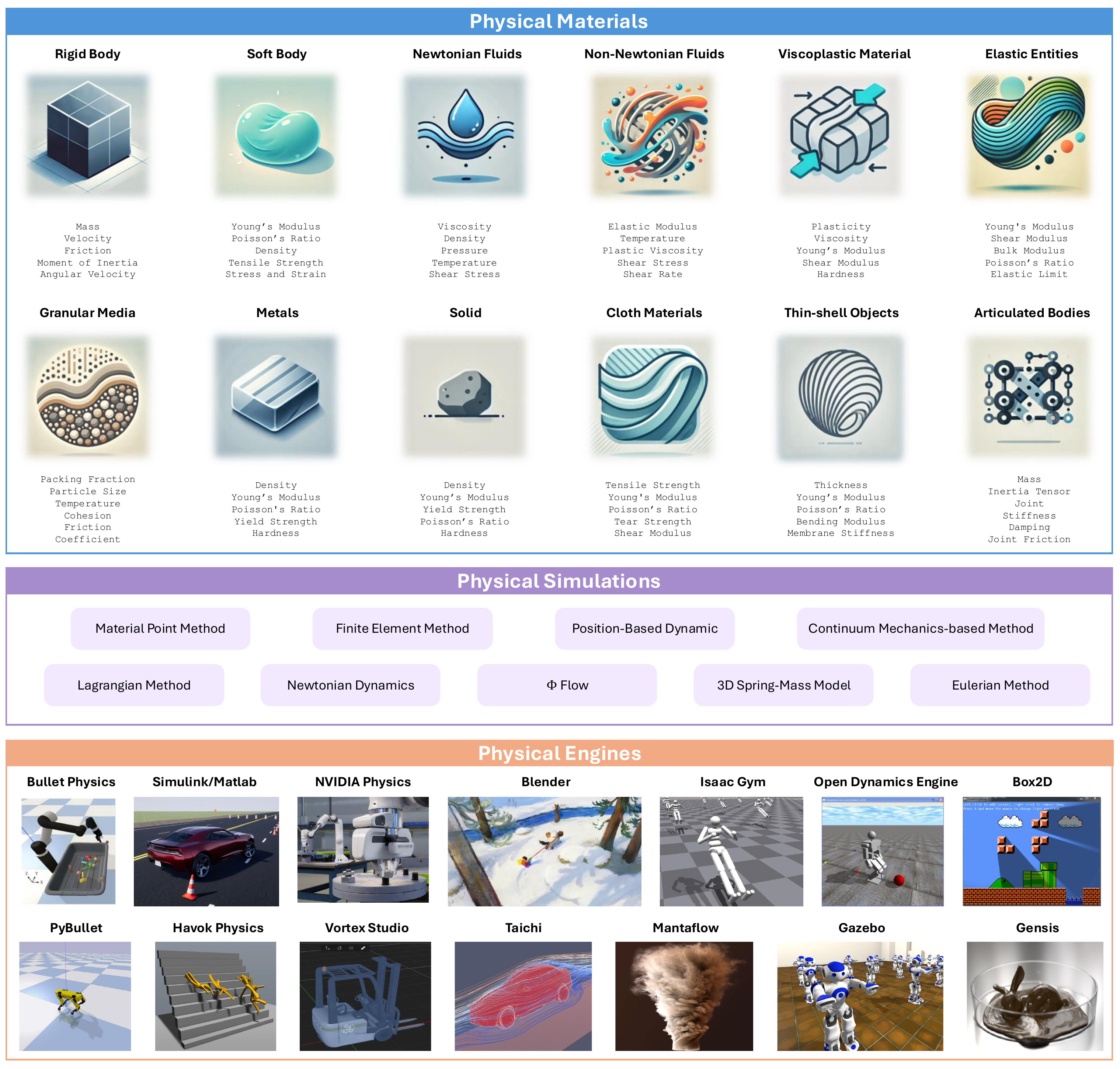}
\end{center}
   \caption{Overview of key components in physical simulation relevant to physics-aware generation, i.e., physical materials, simulation methods, and physical engines. Physical materials define the types of entities being modeled with different assumptions and constraints, with illustrations generated by GPT-4o. Simulation methods provide the computational tools to model the dynamics of these materials under physical laws. Physics engines serve as practical and off-the-shelf platforms implementing these simulations. Examples are given for each component for illustration.}
\label{fig:physics-overview}
\end{figure*}

\section{Physical Simulation}
\label{sec:simulation}
Physical simulation refers to the use of simulated environments or models that mimic real-world physical systems, allowing the generative models to learn and infer without direct interaction with the physical world.
In Fig.~\ref{fig:physics-overview}, we summarize the core elements of physical simulations commonly used in physics-aware generation research, along the dimensions of physical material properties, simulation methods, and off-the-shelf physical engines.
Detailed descriptions are given in the Appendix.


\section{Physical Understanding}
\label{sec:understanding}

Physical understanding in machine learning is the process of inferring the underlying physical models, laws, or parameters from observational data such as images and videos. This capability is crucial for tasks that require modeling real-world dynamics, including object motion, material behavior, and environmental interactions. While extensive research has been dedicated to physical reasoning and model discovery\cite{Cai2024c,Rong2024,Feng2024b,Li2023d,Yu2023c,Guan2022,Chen2022,Jatavallabhula2021,Zhu2024d,Zhai2024,Fu2024,Xu2019,Li2018,Wu2016,Battaglia2016,Davis2017,Guen2020,Ding2021,Mrowca2018,Wu2015,Riochet2022,Andriluka2024,Wu2017,Qiao2022,Cao2024,Kandukuri2022,Jaques2019,Watters2017,Chari2019,Li2020}, this area is largely orthogonal to physics-aware generative modeling, 
and hence outside the core focus of this survey. 
Nonetheless, accurately estimating physical parameters remains a foundational component in enabling physically consistent generation.
In the context of physics-aware generative learning, determining the physical parameters that drive simulations is essential for producing realistic and physically plausible outputs. 
We identify three major approaches to obtain physical parameters as below.

\noindent
\textbf{Manually-Set Physical Parameters.}
In this approach, domain experts explicitly define the physical properties and constraints used in simulations\cite{Feng2024a,Xie2024,Jiang2024a,Barcellona2024,Borycki2024,Lin2024a,Mezghanni2021,Guo2024c,Yan2024,Ni2024a,Chen2024b,Aira2024,Liu2023c,Luo2024,Yang2024e}. These parameters may include material properties (e.g., mass, friction, elasticity), environmental conditions (e.g., gravity, fluid dynamics), and initial configurations. While this method is practical and widely used, it often lacks scalability and adaptability to complex or diverse scenarios.

\noindent
\textbf{Automatically-Learned Physical Parameters.}
Data-driven models can automatically infer physical parameters by learning from visual observations \cite{Li2023b,Kaneko2024,Zhong2024,Zhang2024a,Huang2024b,Liu2024b,Wang2023c,Yuan2023} to avoid inflexible pre-defined parameters. These parameters can be estimated either separately in a standalone stage or optimized jointly with the parameters of the generative model.

\noindent
\textbf{LLM-Reasoned Physical Parameters.}
Recent multimodal large language models (LLMs) enable reasoning about physical systems using both textual and visual information \cite{Zhao2024,Lin2024b,Qiu2024,Mao2024b,Xia2024,Furuta2024,Lv2024,Liu2024a,Tan2024,Hsu2024}. By leveraging contextual knowledge and common sense reasoning, LLMs prompted with descriptions of objects can infer their physical materials and plausible physical configurations for simulation.

\begin{table*}
  \centering
  \begin{tabular}{|p{2.7cm} p{2cm}|p{1.5cm}|p{1.4cm} p{2.5cm}|p{3cm} p{0.7cm} |p{0.7cm}|}
\hline
\textbf{Method} & \textbf{Year} & \textbf{Paradigm} & \textbf{G-Mo} & \textbf{G-Md} & \textbf{P-Mo} & \textbf{P-Pa} & \textbf{P-S} \\
\hline
\cite{Li2023b} PAC-NeRF & ICLR'23 & \textcolor{Cyan}{GnS} & NeRF & Interactive NVS & MPM   & \emojiauto & \emojistar \\
\cite{Kaneko2024} iPAC-NeRF & CVPR'24 & \textcolor{Cyan}{GnS} & NeRF & Interactive NVS & MPM   & \emojiauto & \emojistar \\
\cite{Feng2024a} PIE-NeRF & CVPR'24 & \textcolor{Red}{GtS} & NeRF & Interactive NVS & Q-GMLS/Taichi   & \emojihand & \emojisnow \\
\cite{Xie2024} PhysGaussian & CVPR'24 & \textcolor{Red}{GtS} & GS & Interactive NVS & MPM   & \emojihand & \emojisnow \\
\cite{Zhong2024} Spring-Gau & ECCV'24 & \textcolor{Red}{GtS} & GS & Interactive NVS & 3D Spring-Mass   & \emojiauto & \emojisnow \\
\cite{Jiang2024a} VR-GS & SIGGRAPH'24 & \textcolor{Red}{GtS} & GS & Interactive NVS & XPBD   & \emojihand & \emojisnow \\
\cite{Zhang2024a} PhysDreamer & ECCV'24 & \textcolor{Green}{GcS} & GS, DM & Interactive NVS & MPM   & \emojiauto & \emojisnow \\
\cite{Huang2024b} DreamPhysics & arXiv'24 & \textcolor{Green}{GcS} & GS, DM & Interactive NVS & MPM   & \emojiauto & \emojisnow \\
\cite{Liu2024b} Physics3D & arXiv'24 & \textcolor{Green}{GcS} & GS, DM & Interactive NVS & MPM   & \emojiauto & \emojisnow \\
\cite{liu2024unleashing} PhysFlow & CVPR'25 & \textcolor{Green}{GcS} & GS, DM & Interactive NVS & MPM   & \emojiauto & \emojisnow \\
\cite{Barcellona2024} DreMa & arXiv'24 & \textcolor{Red}{GtS} & GS & Interactive NVS & PyBullet   & \emojihand & \emojisnow \\
\cite{Zhao2024} SimAnything & arXiv'24 & \textcolor{Red}{GtS} & GS & Interactive NVS & MPM   & \emojigpt & \emojisnow \\
\cite{Borycki2024} GASP & arXiv'24 & \textcolor{Red}{GtS} & GS & Interactive NVS  & MPM/Taichi   & \emojihand & \emojisnow \\
\cite{Lin2024a} Phy124 & arXiv'24 & \textcolor{Red}{GtS} & GS & Interactive NVS & MPM   & \emojihand & \emojisnow \\
\cite{Lin2024b} Phys4DGen & arXiv'24 & \textcolor{Red}{GtS} & GS, DM & Interactive NVS & MPM   & \emojigpt & \emojisnow \\
\cite{Qiu2024} FeatureSplatting & ECCV'24 & \textcolor{Red}{GtS} & GS & Interactive NVS & MPM/Taichi   & \emojigpt & \emojisnow \\
\cite{Mao2024b} LIVE-GS & arXiv'24 & \textcolor{Red}{GtS} & GS & Interactive NVS & LLM/Unity/PBD   & \emojigpt & \emojisnow \\
\cite{Xia2024} Video2Game & arXiv'24 & \textcolor{Red}{GtS} & NeRF & Interactive NVS & cannon.js/Blender/Unreal   & \emojigpt & \emojisnow \\
\cite{wang2025decoupledgaussian} DecoupledGS & CVPR'25 & \textcolor{Red}{GtS} & GS & Interactive NVS & MLS-MPM & \emojihand & \emojisnow \\
\cite{Mezghanni2021} Mezghanni et al. & CVPR'21 & \textcolor{Mulberry}{ScG} & GAN & 3D Generation & Bullet   & \emojihand & \emojifire \\
\cite{Wang2023c} DiffuseBot & NeurIPS'23 & \textcolor{Mulberry}{ScG} & DM & 3D Generation & MPM/SoftZoo   & \emojiauto & \emojistar \\
\cite{Guo2024c} PhysComp & arXiv'24 & \textcolor{Mulberry}{ScG} & Other & Image-to-3D & FEM   & \emojihand & \emojifire \\
\cite{Yan2024} PhyCAGE & arXiv'24 & \textcolor{GreenYellow}{SiG} & GS, DM & Image-to-3D & MPM   & \emojihand & \emojisnow \\
\cite{Ni2024a} PhyRecon & NeurIPS'24 & \textcolor{Mulberry}{ScG} & Other & Image-to-3D & IssacGym/DiffTaichi   & \emojihand & \emojistar \\
\cite{Chen2024b} Atlas3D & arXiv'24 & \textcolor{Mulberry}{ScG} & DM & Text-to-3D & Euler/Warp   & \emojihand & \emojifire \\
\cite{Furuta2024} Furuta et al. & arXiv'24 & \textcolor{Mulberry}{ScG} & DM & Text-to-Video & VLMs   & \emojigpt & \emojifire \\
\cite{zhao2025synthetic} Zhao et al. & arXiv'25 & \textcolor{Mulberry}{ScG} & DM & Text-to-Video & Unreal/Blender   & \emojihand & \emojisnow \\
\cite{DSOLi} DSO & arXiv'25 & \textcolor{Mulberry}{ScG} & DM & Image-to-3D & MuJoCo   & \emojihand & \emojifire \\
\cite{Lv2024} GPT4Motion & CVPRW'24 & \textcolor{GreenYellow}{SiG} & DM & Text-to-Video & Blender/LLM   & \emojigpt & \emojisnow \\
\cite{Aira2024} MotionCraft & NeurIPS2024 & \textcolor{GreenYellow}{SiG} & DM & Text-to-Video & $\Phi$-Flow   & \emojihand & \emojisnow \\
\cite{Liu2024a} PhysGen & ECCV'24 & \textcolor{GreenYellow}{SiG} & DM & Image-to-Video & Pymunk   & \emojigpt & \emojisnow \\
\cite{xie2025physanimator} PhysAnimator & CVPR'25 & \textcolor{GreenYellow}{SiG} & DM & Image-to-Video & Taichi   & \emojihand & \emojisnow \\
\cite{Tan2024} PhysMotion & arXiv'24 & \textcolor{Cyan}{GnS} & GS, DM & Image-to-Video & MPM   & \emojigpt & \emojisnow \\
\cite{Hsu2024} AutoVFX & arXiv'24 & \textcolor{GreenYellow}{SiG} & GS & Video-to-Video & Blender   & \emojigpt & \emojisnow \\
\cite{Yuan2023} PhysDiff & ICCV'23 & \textcolor{GreenYellow}{SiG} & DM & Text-to-Motion & IsaacGym   & \emojiauto & \emojisnow \\
\cite{Liu2023c} Liu et al. & ICCV'23 & \textcolor{CadetBlue}{SeG} & Other & Interactive 3D & Customized   & \emojihand & \emojitarget \\
\cite{Luo2024} PhysPart & arXiv'24 & \textcolor{CadetBlue}{SeG} & DM & Interactive 3D & IsaacGym   & \emojihand & \emojitarget \\
\cite{Yang2024e} PhyScene & CVPR'24 & \textcolor{CadetBlue}{SeG} & DM & Interactive 3D Scenes & Omniverse IsaacSim  & \emojihand & \emojitarget \\
\hline

  \end{tabular}
  \caption{Summary of Physics-Aware Generation Methods with Explicit Physical Simulation (PAG-E).\\ \textbf{Paradigm}: Main paradigm of incorporating physics with the generative model {(\textcolor{Red}{GtS}, \textcolor{GreenYellow}{SiG}, \textcolor{Cyan}{GnS}, \textcolor{Mulberry}{ScG}, \textcolor{Green}{GcS}, \textcolor{CadetBlue}{SeG})}. \textbf{G-Mo}: Generative model {(DM, GAN, GS, NeRF, etc.)}. \textbf{G-Md}: Modality of generation, NVS: Novel view synthesis. \textbf{P-Mo}: Physical model. \textbf{P-Pa}: Selection of physical parameters (\emojihand: Manually selected, \emojiauto: Automatically learned, \emojigpt: LLM-reasoned). \textbf{P-S}: Stage of the generative model when incorporating physics {(\emojifire: Train stage, \emojisnow: Testing stage, \emojistar: Train\&Test, \emojitarget: Evaluation/Deployment)}.}
  \label{tab:all_methods}
\end{table*}

\section{Physics-Aware Generation}
\label{sec:generation}

We first provide an overview of the primary classes of generative models employed in physics-aware generation and then introduce the main paradigms for incorporating physics-based constraints and knowledge into these generative models.

\subsection{Generative Models}

We begin by introducing widely used models including Generative Adversarial Networks (GANs), Diffusion Models (DMs), and neural rendering methods including Neural Radiance Field (NeRF) and Gaussian Splatting (GS) which although traditionally viewed as rendering or reconstruction frameworks, are interpreted as generative models in a broader sense in this survey. 
These generative models can handle a variety of data modalities (e.g., images, videos, 3D, 4D, or interactive environments) and encompass various domains (e.g., general data, human-centric data, and indoor scenes).

\subsection{Generative Adversarial Networks}

A Generative Adversarial Network\cite{GAN} consists of two neural networks, i.e., a generator and a discriminator, engaged in a competitive process. The generator seeks to create synthetic data that closely resembles real-world data, while the discriminator attempts to distinguish between real and generated data. This adversarial dynamic drives both networks to improve iteratively, resulting in a generator capable of producing highly realistic data samples.
Specifically, the generator $G$ and the discriminator $D$ play a two-player minimax game with the following loss function:
\begin{equation} \label{eq:gan}
\begin{split}
    \underset{G}{\mathrm{min}} \; \underset{D}{\mathrm{max}} \; \mathcal{L}(G,D) = & \mathbb{E}_{x \sim p_{\mathrm{data}}(x)}[\mathrm{log}D(x)] + \\
    & \mathbb{E}_{z \sim p_{\mathrm{z}}(z)}[\mathrm{log}(1-D(G(z)))],
\end{split}
\end{equation}
where $x$ is a real sample from the data distribution and $G(z)$ is a sample generated using the latent code $z$.
StyleGAN\cite{karras2019style} can provide an automatically learned, unsupervised separation of high-level attributes in latent codes, and later versions further fix characteristic artifacts in high-resolution generation\cite{karras2020analyzing}.
GAN has been the dominant generative model before the popularity of diffusion models.

\subsection{Diffusion Models}

Diffusion Models are a class of generative models that have gained significant attention in recent years for their ability to produce high-quality synthetic data\cite{DiffusionSurvey}.
Compared to GANs, DMs are known for their stability during training but lower efficiency during sampling\cite{BeatsGAN,DDPM}.
Diffusion models leverage a sequential process of transforming simple noise distributions into complex data distributions through a series of learned denoising steps. 
Diffusion models simulate a gradual corruption of data by adding Gaussian noise over multiple steps in a forward process:
\begin{equation} \label{eq:dm_1}
q(x_t|x_{t-1}) = \mathcal{N}(x_t;\sqrt{1-\beta_t}x_{t-1},\beta_t \boldsymbol{I}),
\end{equation}
where $t$ is the diffusion step, $x_t$ is the noisy data at step $t$, $\beta$ is a schedule parameter.
This process is then reversed using a neural network parameterized to denoise the data step-by-step: 
\begin{equation} \label{eq:dm_2}
p_\theta(x_{t-1}|x_{t}) = \mathcal{N}(x_{t-1};\boldsymbol{\mu}_\theta(x_{t},t), \sigma_t^2 \boldsymbol{I}),
\end{equation}
where $\sigma$ is the variance schedule.
This denoising process can be parameterized in several ways, including mean prediction, noise prediction, and clean data prediction\cite{luo2022understanding}.
This iterative refinement allows diffusion models to generate detailed and diverse outputs while maintaining strong theoretical foundations.
Score-based models are of an equivalent formulation as diffusion models but from the differential equation perspective\cite{score1,score2,score3}.
The denoising process can be accelerated by samplers such as DDIM\cite{DDIM} and DPM-Solvers\cite{lu2022dpm}.
Latent diffusion models\cite{LDM} conduct the forward and reverse processes in a high-level latent space and delegate low-level reconstruction to an auxiliary VAE.
Techniques like classifier guidance and classifier-free guidance further improve controllability and generation quality\cite{CFG}.
Diffusion model and its later developments\cite{Peebles2022,Zhang2023,Lipman2023,Liu2023b,Song2023,Meng2022,Karras2022} have greatly advanced generative tasks for visual data.

\subsection{Neural Radiance Field}

Neural Radiance Field\cite{gao2022nerf} is an approach to scene representation and novel view synthesis, which leverages neural networks as implicit representation to model the volumetric properties of a scene. 
NeRF utilizes a multi-layer perceptron (MLP) to encode a continuous volumetric scene as a function mapping 3D spatial coordinates $x, y, z$ and viewing directions $\theta, \phi$ to radiance density $\sigma$ and color $c = (r, g, b)$ values:
\begin{equation} \label{eq:nerf}
c, \sigma = F(x, y, z, \theta, \phi).
\end{equation}
By optimizing the network to minimize the error in rendering the observed views of a scene, NeRF can synthesize photorealistic novel views by querying the MLP along camera rays and using volume rendering techniques to project the output colors and densities into an image.
PixelNeRF\cite{PixelNeRF} enables neural scene representation from only one single observed image.
MIP-NeRF\cite{MipNerf} reduces aliasing artifacts and significantly improves the ability to represent fine details.
DNeRF\cite{DNerf} extends neural radiance fields to dynamic objects with motions.
These advancements have propelled the neural radiance field into widespread applications in physics-informed scene representations.

\subsection{Gaussian Splatting}

Gaussian Splatting\cite{Fei2024} is an approach to scene reconstruction and novel view synthesis that represents a scene as a set of dense, overlapping Gaussian “blobs” in 3D space with learned parameters including the mean $\mu_i$, covariance $\Sigma_i$, opacity $\alpha_i$, and view-dependent color $c_i(\theta, \phi)$:
\begin{equation} \label{eq:gs}
\mathcal{G}(x, y, z, \mu_i, \Sigma_i, \alpha_i) \cdot c_i(\theta, \phi).
\end{equation}
Instead of relying on volumetric grids or implicit neural fields, Gaussian splatting is an explicit radiance field-based scene representation.
During rendering, the Gaussians are “splatted” onto the image plane, with their contributions composited together to form a final image. 
By jointly optimizing the positions, shapes, and colors of the Gaussians, Gaussian splatting achieves photorealistic renderings with compelling speed and accuracy, making it an attractive alternative or complement to grid-based and neural implicit methods like NeRF.
4D Gaussian Splatting (4D-GS)\cite{Kim2024} goes beyond static scenes and extends to dynamic scenes as a 4D representation.
SplatterImage\cite{SplatterImage} maps a single input image to one 3D Gaussian per pixel, enabling fast scene reconstruction from a single view.
LGM\cite{tang2025lgm} is proposed for 3D generation, which generates 3D Gaussians from multi-view images synthesized by external models at inference time from text or single-image inputs

\subsection{Physics-Aware Generation w/ Explicit Simulation}

In the following, we review papers working on physics-aware generation with explicit physical simulation (PAG-E), which are grouped into six categories (Fig.~\ref{fig:tree-structure1}) according to the paradigm of how physics simulation (Sim) is integrated with the generative model (Gen).
Note that one paper 
may follow multiple paradigms, and we assign it to the most relevant one in such cases.



     
\definecolor{FillColor}{RGB}{255,255,242}  

\begin{figure*}
\centering
\resizebox{\textwidth}{!}
{
\begin{tikzpicture}[
    scale=1, %
    transform shape, %
    edge from parent fork right, grow=right, 
    level 1/.style={
        text width=5cm,
        level distance=3cm %
    },
    level 2/.style={
        text width=9cm,
        level distance=7cm %
    }, 
    level 3/.style={
        text width=9cm,
        level distance=10cm %
    }, 
    every node/.style={
        draw, 
        rectangle, 
        rounded corners = 3pt, 
        align=center,  
        inner sep=5pt, %
        font=\large %
    } %
]
    \node [rotate=90]{\textbf{Physics-Aware Generation with Explicit Simulation}}
    child {node[yshift=14.8cm, xshift=2cm, draw=black] {Paradigm 1 (Sec.\ref{sec:para1}) \\ \textbf{Gen-to-Sim} }
        child {node[xshift=2cm, draw=black] {Applications in VR and Robotics}
            child{node[fill=FillColor][anchor=west][minimum height=3.3em][xshift=-4.5cm]{VR-GS\cite{Jiang2024a}, LIVE-GS\cite{Mao2024b}, DreMa\cite{Barcellona2024}
            }}}
        child {node[xshift=2cm, draw=black] {Physical Feature Field}
            child{node[fill=FillColor][anchor=west][minimum height=3.3em][xshift=-4.5cm]{Feature Splatting\cite{Qiu2024}, Phys4DGen\cite{Lin2024b}, SimAnything\cite{Zhao2024}
            }}}
        child {node[xshift=2cm, draw=black] {Gaussian Blobs as Simulation Elements}
            child{node[fill=FillColor][anchor=west][minimum height=3.3em][xshift=-4.5cm]{PhysGaussian\cite{Xie2024}, GASP\cite{Borycki2024}, Spring-Gau\cite{Zhong2024}, Phy124\cite{Lin2024a}, DecoupledGS~\cite{wang2025decoupledgaussian}
            }}}
        child {node[xshift=2cm, draw=black] {Simulation Elements in NeRF}
            child{node[fill=FillColor][anchor=west][minimum height=3.3em][xshift=-4.5cm]{PIE-NeRF\cite{Feng2024a}, Video2Game\cite{Xia2024}
            }}}
    }
    child {node[yshift=7.9cm, xshift=2cm, draw=black] {Paradigm 2 (Sec.\ref{sec:para2}) \\ \textbf{Sim-in-Gen} }
        child {node[xshift=2cm, draw=black] {Generative Programs for Simulation}
            child{node[fill=FillColor][anchor=west][minimum height=3.3em][xshift=-4.5cm]{AutoVFX\cite{Hsu2024}, GPT4Motion\cite{Lv2024}
            }}}
        child {node[xshift=2cm, draw=black] {Simulation as Optimization}
            child{node[fill=FillColor][anchor=west][minimum height=3.3em][xshift=-4.5cm]{PhyCAGE\cite{Yan2024}, PhysDiff\cite{Yuan2023}
            }}}
        child {node[xshift=2cm, draw=black] {Conditions from Simulation}
            child{node[fill=FillColor][anchor=west][minimum height=3.3em][xshift=-4.5cm]{GPT4Motion\cite{Lv2024}, MotionCraft\cite{Aira2024}, PhysGen\cite{Liu2024a}, PhysAnimator~\cite{xie2025physanimator}
            }}}
    }
    child {node[yshift=2.5cm, xshift=2cm, draw=black] {Paradigm 3 (Sec.\ref{sec:para3}) \\ \textbf{Gen-and-Sim} }
        child {node[xshift=2cm, draw=black] {Alternating Simulation and Generation}
            child{node[fill=FillColor][anchor=west][minimum height=3.3em][xshift=-4.5cm]{PhysMotion\cite{Tan2024}
            }}}
        child {node[xshift=2cm, draw=black] {Joint Geometry and Physics Learning}
            child{node[fill=FillColor][anchor=west][minimum height=3.3em][xshift=-4.5cm]{PAC-NeRF\cite{Li2023b}, iPAC-NeRF\cite{Kaneko2024}
            }}}
    }
    child {node[yshift=-3.5cm, xshift=2cm, draw=black] {Paradigm 4 (Sec.\ref{sec:para4}) \\ \textbf{Sim-Constrained Gen} }
        child {node[xshift=2cm, draw=black] {Training Reference Model using Simulation Data}
            child{node[fill=FillColor][anchor=west][minimum height=3.3em][xshift=-4.5cm]{Zhao et al.~\cite{zhao2025synthetic}
            }}}
        child {node[xshift=2cm, draw=black] {Post-training with Physical Feedback}
            child{node[fill=FillColor][anchor=west][minimum height=3.3em][xshift=-4.5cm]{Furuta et al.\cite{Furuta2024}, DSO~\cite{DSOLi}
            }}}
        child {node[xshift=2cm, draw=black] {Simulation for Data Filtering}
            child{node[fill=FillColor][anchor=west][minimum height=3.3em][xshift=-4.5cm]{DiffuseBot\cite{Wang2023c}
            }}}
        child {node[xshift=2cm, draw=black] {Loss Functions from Simulation}
            child{node[fill=FillColor][anchor=west][minimum height=3.3em][xshift=-4.5cm]{PhysComp\cite{Guo2024c}, PhyRecon\cite{Ni2024a}, Atlas3D\cite{Chen2024b}, Mezghanni et al.\cite{Mezghanni2021}
            }}}
    }
    child {node[yshift=-9.5cm, xshift=2cm, draw=black] {Paradigm 5 (Sec.\ref{sec:para5}) \\ \textbf{Gen-Constrained Sim} }
        child {node[xshift=2cm, draw=black] {Learning Physical Parameters from Generated Data}
            child{node[fill=FillColor][anchor=west][minimum height=3.3em][xshift=-4.5cm]{PhysDreamer\cite{Zhang2024a}
            }}}
        child {node[xshift=2cm, draw=black] {Score Distillation Sampling}
            child{node[fill=FillColor][anchor=west][minimum height=3.3em][xshift=-4.5cm]{Physics3D\cite{Liu2024b}, DreamPhysics\cite{Huang2024b}, PhysFlow~\cite{liu2024unleashing}
            }}}
    }
    child {node[yshift=-13.5cm, xshift=2cm, draw=black] {Paradigm 6 (Sec.\ref{sec:para6}) \\ \textbf{Sim-evaluated Gen} }
            child{node[fill=FillColor][anchor=west][minimum height=3.3em][xshift=-3cm]{Liu et al.\cite{Liu2023c}, PhysPart\cite{Luo2024}, PhyScene\cite{Yang2024e}
            }}
    };
\end{tikzpicture}
}
\caption{Structure of Paradigms, Ideas, and Methods in Physics-Aware Generation with Explicit Physical Simulation.}
\label{fig:tree-structure1}
\end{figure*}

\subsubsection{Paradigm 1: Gen-to-Sim (GtS)}
\label{sec:para1}

This category of 
methods usually appends physical properties to a generative representation in a \textit{post-processing} manner to make it simulatable and interactable.

\noindent
\textbf{Simulation Elements in NeRF.}
PIE-NeRF\cite{Feng2024a} uses Poisson disk sampling to distribute ``particles" in the NeRF density field, which form simulation elements through Voronoi grouping. 
Quadratic generalized moving least square (Q-GMLS) strategy and Lagrangian dynamics are then applied to the elements for simulation, enabling users to interact with the NeRF scene using external forces.
Video2Game\cite{Xia2024} automatically converts a single video of a real-world scene into an interactive virtual environment.
To achieve this, a NeRF-generated scene is segmented into individual objects. 
Each segmented object is then assigned physical parameters such as mass, friction, and collision geometry (e.g., sphere, box, convex polygon) to be simulated with rigid-body physics in real time using WebGL-based game engine.

\noindent
\textbf{Gaussian Blobs as Simulation Elements.}
PhysGaussian\cite{Xie2024} first builds a Gaussian Splatting with anisotropic regularization, and then imparts physically grounded behavior to the Gaussian kernels through Material Point Method (MPM). 
The method applies continuum mechanics to evolve 3D Gaussian kernels, where the kernels are treated as discrete particle clouds for discretizing the continuum.
This allows the system to model the deformation realistically by tracking physical quantities like stress and strain.
Similarly, GASP\cite{Borycki2024} also leverages the Material Point Method (MPM) to simulate physical behaviors.
GASP converts Gaussian components into triangle-based meshes, on which the MPM is applied, and the results are re-parameterized back into Gaussian components for rendering. 
Spring-Gau\cite{Zhong2024} first reconstructs static Gaussian, from which anchor points are sampled and are connected via ``springs" to model elastic behavior.
The method uses differentiable simulation to optimize learnable stiffness and damping parameters from video observations, to transform the Gaussian representation to be dynamic and simulatable.
Phy124\cite{Lin2024a} generates physics-consistent 4D content from a single image, by evolving dynamic 3D content adhering to natural physical laws.
Specifically, 3D Gaussians are generated from a single image using diffusion prior and then animated by attaching an MPM simulator.
DecoupledGS~\cite{wang2025decoupledgaussian} separates objects and contact surfaces during scene restoration to improve the realism of interactive physical simulations such as user-specified external forces, scene collisions, and object fracturing.

\noindent
\textbf{Physical Feature Field.}
Feature Splatting\cite{Qiu2024} extends Gaussian Splatting by embedding semantic features extracted from vision-language models.
Such feature-carrying 3D Gaussians bridge the gap between static 3D scene representations and dynamic physical behaviors.
The physics engine simulates physical interactions by treating Gaussian centroids as particles assigned with material properties identified by natural language semantics.
Phys4DGen\cite{Lin2024b} utilizes the Segment Anything model and large language models to infer the material composition and physical properties of different object components in a scene represented by Gaussian Splatting. 
This allows for more accurate physical simulation by assigning realistic material behaviors to different object parts.
SimAnything\cite{Zhao2024} employs a multi-modal large language model to predict the mean physical properties at the object level, which further helps the estimation of the probability distribution of physical properties at a particle level for MPM simulation.

\noindent
\textbf{Applications in VR and Robotics.}
VR-GS\cite{Jiang2024a} introduces a framework designed for interactive and physically realistic manipulation of 3D content in Virtual Reality (VR), which integrates an XPBD physics simulation model with 3D Gaussian Splatting to support real-time and physics-aware interaction.
LIVE-GS\cite{Mao2024b} further incorporates GPT-4 to analyze and infer the physical properties of objects directly from images, bypassing the manual tuning in VR-GS.
DreMa\cite{Barcellona2024} combines scene reconstruction with physics-based simulation to create a manipulable, object-centric world model for robotics, enabling robots to imagine new object configurations and predict the outcomes of their actions.


\subsubsection{Paradigm 2: Sim-in-Gen (SiG)}
\label{sec:para2}

This paradigm features physics simulation integrated directly into the generative model, functioning as a core sub-module.

\noindent
\textbf{Conditions from Simulation.}
In GPT4Motion\cite{Lv2024}, Blender-based physics simulation is integrated into the video generation pipeline. 
Specifically, GPT-4 translates user prompts into Python scripts that control Blender’s physics engine to simulate realistic motion. The output of Blender, rendered as edge and depth maps, is then fed into ControlNet-modified Stable Diffusion to generate videos that are both visually consistent and physically plausible. 
MotionCraft\cite{Aira2024} enhances pretrained image diffusion models and introduces motions by warping the noise latent space with optical flow derived from physics simulation. This process allows the model to generate temporally consistent frames that evolve according to physically accurate dynamics such as fluid dynamics, rigid body motion, and multi-agent interactions.
Similarly, by using image-based warping with simulated motion dynamics, PhysGen\cite{Liu2024a} generates physically plausible video outputs given an input image and user-defined forces and torques.
The simulation is based on rigid-body dynamics governed by Newton’s Laws, using physical parameters inferred by large foundation models.
PhysAnimator~\cite{xie2025physanimator} further extends to deformable body dynamics by using mesh reconstruction and simulation to warp a sketch image.
The warped sketches are then sent into ControlNet to guide the video diffusion model for generation.


\noindent
\textbf{Simulation as Optimization.}
PhyCAGE\cite{Yan2024} treats MPM simulation as a physics-aware optimizer by evolving the particle system based on physical laws.
The gradient of the loss function is interpreted as the initial velocity of the particles derived from 3D Gaussians, and this velocity is passed into the MPM simulation to optimize the system over sub-steps before each gradient descent step.
PhysDiff\cite{Yuan2023} generates physically plausible human motions by integrating physics-based constraints directly into the sampling process of diffusion models.
At each denoising step, the intermediate motion is passed through a physics simulator for correction.
The physically-corrected motion then feeds back into the sampling process and guides subsequent denoising steps, steering the generation toward physically plausible motions.

\noindent
\textbf{Generative Programs for Simulation.}
AutoVFX\cite{Hsu2024} works on photorealistic and physically plausible video editing through natural language instructions, which combines LLM-based code generation and physics-based simulation.
User instructions are transformed into executable code, enabling edits such as object insertion, material changes, dynamic interactions, and particle effects.
The code is executed in the Blender engine to render the scene with photorealistic lighting and materials to produce the edited video.
Similarly, GPT4Motion\cite{Lv2024} uses large language models to automate the generation of Blender Python scripts that drive physical simulations, integrating semantic understanding with procedural generation to create complex and physics-based animations to guide video generation.

\subsubsection{Paradigm 3: Gen-and-Sim (GnS)}
\label{sec:para3}

This paradigm involves simultaneous or interconnected operation of generation and simulation processes with tight coupling.

\noindent
\textbf{Joint Geometry and Physics Learning.}
PAC-NeRF\cite{Li2023b} addresses the challenge of inferring both the geometric and physical parameters of objects solely from multi-view video data, using a hybrid Eulerian-Lagrangian representation of the scene.
Concretely, the Eulerian grid representation is used for NeRF to learn the geometry, while the Lagrangian particle representation is used for simulation to learn the physical parameters.
iPAC-NeRF\cite{Kaneko2024} further proposes Lagrangian Particle Optimization to directly optimize the positions and features of particles in the Lagrangian space, leading to dynamic refinement of the geometric structure across the entire video sequence while adhering to physical constraints.

\noindent
\textbf{Alternating Simulation and Generation.}
PhysMotion\cite{Tan2024} follows a generation-simulation-generation process for image-to-video generation.
Firstly, the foreground object in the input image is converted into a coarse 3D Gaussian Splatting representation.
Then, the model undergoes simulation using MPM, applying physical laws to simulate how the object would behave under forces, to produce a coarse video depicting physics-grounded dynamics.
Finally, the coarse video is further refined using diffusion-based video enhancement to improve visual realism.

\subsubsection{Paradigm 4: Sim-Constrained Gen (ScG)}
\label{sec:para4}

This is a paradigm where the simulation imposes constraints or guidance on the training of generative models to improve physical awareness.

\noindent
\textbf{Loss Functions from Simulation.}
PhysComp\cite{Guo2024c} creates 3D models from single images while ensuring physical compatibility.
It is constrained by a simulation-based physical model—specifically, the static equilibrium constraint—which ensures the generated 3D shapes behave realistically under physical forces.
PhyRecon\cite{Ni2024a} incorporates simulation as loss functions to constrain and guide the generation process toward physically plausible 3D scenes. 
This is achieved through the integration of a differentiable particle-based physical simulator, which directly influences the neural implicit surface representation to improve stability and model physical uncertainty.
Atlas3D\cite{Chen2024b} generates 3D models that are self-supporting, incorporating a standability loss to penalize rotational instability to ensure the generated model maintains its upright orientation during simulation, and a stable equilibrium loss to encourage resilience against minor perturbations.
Mezghanni et al.\cite{Mezghanni2021} introduce two differentiable physical loss functions, i.e., a connectivity loss to regularize that the generated 3D shapes are single connected components, and a stability loss to promote physical stability under gravity.

\noindent
\textbf{Simulation for Data Filtering.}
DiffuseBot\cite{Wang2023c} employs simulation as a component for data filtering during its embedding optimization phase.
Specifically, 3D robot models generated from diffusion models are evaluated using a differentiable physics simulation to assess the performance of each robot based on task-specific metrics.
A filtering mechanism then selects the top-performing designs to be retained, gradually skewing the sampling distribution of the generative model toward successful robot designs.

\noindent
\textbf{Post-training with Physical Feedback.}
Furuta et al.\cite{Furuta2024} addresses the challenge of generating realistic dynamic object interactions in text-to-video models, by using reinforcement learning (RL) fine-tuning from external feedback, particularly from vision-language models (VLMs).
The VLM-based feedback mechanism serves as a form of simulation or evaluation of physical realism, which constrains and guides the generation model toward producing physically plausible videos.
DSO~\cite{DSOLi} creates 3D assets using a base image-to-3D generation model and annotates their self-standing stability through physics-based simulation. The base model is then fine-tuned on the created dataset using either Direct Preference Optimization (DPO) or Direct Reward Optimization (DRO) to encourage stable samples.

\noindent
\textbf{Training Reference Model using Simulation Data.}
Zhao et al.~\cite{zhao2025synthetic} improve video generation by introducing a reference model, which is an extra video generation model in parallel with the primary generation model, but trained on synthetic data from simulation engines.
The authors propose a strategy called SimDrop, where the auxiliary reference model guides the primary model to minimize artifacts unique to synthetic data while retaining the desired physical fidelity.
To achieve this, a weighted combination of the primary model's output and the reference model's output is used in a manner similar to the classifier-free guidance of diffusion models during inference.

\subsubsection{Paradigm 5: Gen-Constrained Sim (GcS)}
\label{sec:para5}

In this paradigm, the generation model acts as a guidance or prior knowledge for the simulation process.

\noindent
\textbf{Score Distillation Sampling.}
Physics3D\cite{Liu2024b} combines elastoplastic and viscoelastic MPM with a video diffusion model.
To optimize physical parameters, it utilizes Score Distillation Sampling (SDS) from a pretrained video diffusion model. 
The diffusion model provides priors learned from video data, guiding the simulation toward physically plausible behaviors without requiring manual specification of material properties.
DreamPhysics\cite{Huang2024b} further proposes Motion Distillation Sampling (MDS) to better capture motion-specific priors and reduce color bias in the optimization.
Similarly, PhysFlow~\cite{liu2024unleashing} utilizes optical flow loss to distil the knowledge from video diffusion models to optimize the material properties intialized by multimodal foundation models.

\noindent
\textbf{Learning Physical Parameters from Generated Data.}
PhysDreamer\cite{Zhang2024a} synthesizes realistic interactive dynamics for objects by extracting the priors of physical dynamics from video generation models. Different from score distillation sampling, this method learns such priors from generated videos.
The method first uses a pre-trained image-to-video generation model to create a reference video of a static object undergoing plausible motions.
Then it optimizes physical parameters such as Young's modulus in MPM simulation by maximizing the visual similarity between the simulation-rendered video and the generated reference video.
 
\subsubsection{Paradigm 6: Sim-evaluated Gen (SeG)}
\label{sec:para6}

In this paradigm, generation is intended for simulation-based deployment, focusing on the utility in simulated environments.

Liu et al.\cite{Liu2023c} generates physically plausible articulated 3D assets from limited data.
By using auxiliary penetrations penalization and collision-based shape optimization, this method produces articulated meshes that behave correctly under simulation.
PhysPart\cite{Luo2024} generates 3D replacement parts that fit precisely and move smoothly, essential for applications such as 3D printing and robotic manipulation.
The approach integrates geometric conditioning through classifier-free guidance and enforces physical constraints via stability and mobility losses during the sampling process.
PhyScene\cite{Yang2024e} generates high-quality interactive 3D scenes tailored for embodied artificial intelligence, which leverages a conditional diffusion model to capture scene layouts and employs physics-based guidance functions, ensuring the scene is physically functional and suitable for embodied agent interactions.


\definecolor{FillColor}{RGB}{255,255,242}  

\begin{figure*}
\centering
\resizebox{\textwidth}{!}
{
\begin{tikzpicture}[
    scale=1, %
    transform shape, %
    edge from parent fork right, grow=right, 
    level 1/.style={
        text width=10cm,
        level distance=3.5cm %
    },
    level 2/.style={
        text width=9cm,
        level distance=11cm %
    }, 
    level 3/.style={
        text width=9cm,
        level distance=12cm %
    }, 
    every node/.style={
        draw, 
        rectangle, 
        rounded corners = 3pt, 
        align=center,  
        inner sep=5pt, %
        font=\large %
    } %
]
    \node [rotate=90, text width=6cm]{\textbf{Physics-Aware Generation w/o Explicit Simulation}}
    child {node[yshift=6.5cm, xshift=4cm, draw=black] {Physical Awareness Emergent in Large Video Models}
            child{node[fill=FillColor][anchor=west][minimum height=3.3em][xshift=-4cm]{Sora~\cite{Sora}, OpenSora~\cite{OpenSora}, CogVideoX~\cite{CogvideoX}, ModelScope~\cite{ModelScope}, Cosmos~\cite{Cosmos}, etc.}}
    }
    child {node[yshift=3.5cm, xshift=4cm, draw=black] {Physical Awareness from Large Language Models}
            child{node[fill=FillColor][anchor=west][minimum height=3.3em][xshift=-4cm]{PhyT2V\cite{PhyT2V}, VideoAgent~\cite{soni2024videoagent}, etc.
            }}
    }
    child {node[yshift=0.5cm, xshift=4cm, draw=black] {Physical Awareness from Physics-rich Training Data}
            child{node[fill=FillColor][anchor=west][minimum height=3.3em][xshift=-4cm]{WISA~\cite{wang2025wisa}, PISA~\cite{li2025pisa}, etc.
            }}
    }
    child {node[yshift=-3.0cm, xshift=4cm, draw=black] {Generative Interactive Dynamics and Motion Controls}
            child{node[fill=FillColor][anchor=west][minimum height=3.3em][xshift=-4cm]{Blattmann et al.\cite{Blattmann2021a}, Generative Image Dynamics\cite{Li2023a}, Motion Prompting\cite{Geng2024b}, VideoComposer\cite{Wang2023b}, Yoda\cite{Davtyan2024}, Motion Dreamer\cite{Xu2024b}, Motion Guidance\cite{Geng2024a}, LivePhoto\cite{Chen2023a}, etc.
            }}
    }
    child {node[yshift=-6.5cm, xshift=4cm, draw=black] {Physical Domain Data Generation}
            child{node[fill=FillColor][anchor=west][minimum height=3.3em][xshift=-4cm]{CoCoGen \cite{Jacobsen2023}, Cao et al.\cite{cao2024teaching}, etc.
            }}
    };
\end{tikzpicture}
}
\caption{Structure of Ideas and Methods in Physics-Aware Generation without Explicit Physical Simulation.}
\label{fig:tree-structure2}
\end{figure*}

\subsection{Physics-Aware Generation w/o Explicit Simulation}

In this section, we introduce representative works of generative learning in vision that exhibit a degree of physical awareness without relying explicitly on physical simulation (PAG-I).

\noindent
\textbf{Physical Awareness Emergent in Large Video Models.}
Recent advancements in large-scale video generation models, such as Sora~\cite{Sora}, OpenSora~\cite{OpenSora}, CogVideoX~\cite{CogvideoX}, VideoCrafter~\cite{Chen2023b}, LAVIE~\cite{Wang2023a}, ModelScope~\cite{ModelScope}, and Cosmos~\cite{Cosmos} demonstrate emergent physical reasoning capabilities. 
These models, scaled up and trained on vast datasets of internet-scale video content, have shown promising abilities to implicitly capture and replicate certain physical dynamics and causal relationships present in the real world~\cite{Bansal2024}. 
This emergent physical understanding enables them to generate complex scenes with coherent object interactions, realistic motions, and plausible temporal dynamics.
This progress positions them as potential foundations for universal simulators capable of modeling diverse physical phenomena, despite the lack of of explicitly programmed physics engines\cite{Zhu2024a}.
However, this implicit physical reasoning remains in its early stage and systematic evaluations are still open areas. 
For example, PhyGenBench\cite{Meng2024} introduces a comprehensive benchmark designed to evaluate the physical common sense understanding of text-to-video generation models.  
The findings highlight that current models, despite their impressive visual generation capabilities, struggle to represent basic physical laws accurately.
Kang et al.\cite{Kang2024} investigate whether large-scale video generation models can autonomously learn and generalize physical laws solely from visual data.
It is found that scaling up the model size and dataset does not help with out-of-distribution physical generalization.
Current models primarily rely on the existence of closely similar training examples rather than abstracting general physical rules, underscoring the need for more targeted methods to develop fully physics-aware generative models.

Cosmos Platform~\cite{Cosmos} is a recent open-source toolkit, including a video data pipeline, video tokenizers, pre-trained models, and post-trained models.
Transformer-based diffusion models and transformer-based autoregressive models are trained on massive video datasets as the pre-trained world foundation models, which can be further fine-tuned on datasets for specific tasks demanding physical awareness such as robotic manipulation, camera control, and autonomous driving.
Cosmos-Reason1~\cite{Cosmos-Reason1} focuses on developing multimodal large language models for embodied decisions, integrating diverse training strategies of supervised fine-tuning and reinforcement learning in terms of physical common sense, intuitive physics, and embodied reasoning.
Cosmos-Reason1 also defines an ontology for physical common sense with three main categories (Space,
Time, and Fundamental Physics) and 16 fine-grained categories.
Cosmos-Transfer1~\cite{Cosmos-Transfer1} further enables conditional world generation controlled by different spatial inputs such as segmentation, depth, and edge maps, using a ControlNet-like architecture for adaptive multimodal control.

\noindent
\textbf{Physical Awareness from Large Language Models.}
Large language models can also supply physical knowledge for improving the generation of visual data.
PhyT2V\cite{PhyT2V} addresses the lack of physical realism in text-to-video models by integrating large language models to refine textual prompts iteratively.
The process first extracts objects and relevant physical rules from the initial text prompt, and then uses a video captioning model to generate semantic descriptions of the generated video and compare it with the original prompt to identify mismatches.
Finally, the text prompt is refined based on the extracted physical properties to correct the mismatches.
VideoAgent~\cite{soni2024videoagent} produces visual plans to guide robotic systems. 
It refines generated video plans by leveraging self-conditioning consistency and feedback from a pretrained vision-language model. 
Additionally, ongoing feedback can be gathered from the environment to enable the continuous improvement of the video generation.

\noindent
\textbf{Physical Awareness from Physics-rich Training Data.}
Intentionally collected domain datasets with rich physical information can be resources for learning physical principles.
WISA~\cite{wang2025wisa} collects the WISA-32K dataset containing approximately 32,000 videos showcasing 17 distinct physical phenomena. 
The dataset includes physics events across three primary domains, i.e., dynamics, thermodynamics, and optics, with detailed physical descriptions and attributes.
The collected data is used to train a video generation model incorporated with a physical property embedding, a mixture-of-physical-experts attention, and a physical classifier.
PISA~\cite{li2025pisa} collected a dataset of real-world videos (361 videos) and synthetic videos (60 videos) showing objects dropping in diverse environments.
The data is employed to improve the physical accuracy of video generation models through post-training techniques, including supervised fine-tuning and object reward optimization that encourages the alignment in terms of segmentation, optical flow, and depth maps.

\noindent
\textbf{Generative Interactive Dynamics and Motion Controls.}
Generative interactive dynamics focus on modeling the natural dynamics of objects without requiring explicit physical simulations, implicitly learning the physical relationships and interactions.
Blattmann et al.\cite{Blattmann2021a} makes an early attempt to predict object dynamics in response to localized user interactions within an image-to-video formulation.
Generative Image Dynamics\cite{Li2023a} generates realistic scene motion from a single image, via a latent diffusion model that predicts spectral volumes representing dense long-term pixel trajectories.
Motion-controlled video generation and video editing belong to another form of generative dynamics\cite{Chen2023a,Chen2023d}.
Motion Prompting\cite{Geng2024b} leverages motion trajectories as conditioning signals, allowing for fine-grained and flexible control of local and global scene motions.
Similarly, VideoComposer\cite{Wang2023b} and Yoda\cite{Davtyan2024} utilize motion vectors as explicit temporal control signals to allow for precise capture of inter-frame dynamics and patterns.
Motion Dreamer\cite{Xu2024b} decouples motion reasoning from high-fidelity video generation, using intermediate motion representations such as optical flow, instance segmentation maps, and depth maps.
Motion Guidance\cite{Geng2024a} controls the diffusion generation process using gradients derived from optical flow estimators, enabling detailed motion edits without model retraining.

\noindent
\textbf{Physical Domain Data Generation.}
CoCoGen \cite{Jacobsen2023} models domain-specific data in physics, such as Darcy Flow and Burgers Equation, using score-based generative models. 
The method enforces physical consistency by injecting discretized partial differential equation information directly into the sampling process of score-based models.
Cao et al.\cite{cao2024teaching} enhance video diffusion models with latent physical knowledge for generating videos that require high physical accuracy, such as fluid simulation and typhoon phenomenon videos.
The method pretrains masked autoencoders on physical phenomena data to capture latent physical knowledge, which is then projected into a quaternion space to construct pseudo-language prompts that guide the video diffusion model in creating physically plausible outputs.

\section{Physics Evaluation}
\label{sec:evaluation}

In this section, we introduce the benchmarks and metrics designed for evaluating the physics awareness of image or video generation models.
Conventionally, most of the image and video generation methods are evaluated based on Frechet Inception Distance (FID) \cite{lucic2018gans}, CLIP Similarity \cite{hessel2021clipscore},  CLIP-FID \cite{kynkaanniemi2022role}, and Inception Score (IS) \cite{lucic2018gans}. 
However, there are a number of gaps in these metrics since they are biased toward the alignment in terms of visual contents and textual semantics. 
For video generation, the detection of unrealistic patterns violating physical laws is more important, which cannot be detected by these metrics effectively. 
Furthermore, the evaluation of Fréchet Video Distance (FVD) requires a reference video dataset, which is difficult to obtain for newly generated scenes. 
These challenges prevent these metrics from effectively evaluating physical commonsense. Some works use Vision Language Models (VLMs) as general evaluators \cite{he2024videoscore, sun2024t2v}, yet these works focus on the evaluation of spatial relationships. Most of them face challenges when generalizing to physical correctness.


\subsection{Benchmarks}



\noindent
\subsubsection{Text-Conditioned Benchmarks}
Text-conditioned benchmarks evaluate the generated images or videos based on the physical phenomenon described in the prompts. 

\noindent
\textbf{PhyBench.}
PhyBench~\cite{meng2024phybench} is a benchmark specifically designed to evaluate the physical commonsense reasoning capabilities of text-to-image models.
It defines physical commonsense as four types, including mechanics, optics, thermal,, and material properties.
It comprises 700 prompts spanning 31 physical scenarios, where the benchmark is constructed by extending designed prompts describing different types of physical scenarios. For example, given the \textit{``pressure"} topic, the authors design a prompt \textit{``an empty plastic bottle at the bottom of the sea"}. 
After that, \textit{``empty plastic bottle"} will be replaced by other objects, and the scenario \textit{``at the bottom of the sea"} is also replaced by other scenarios to describe other topics. 
PhyBench introduces an automated evaluation framework called PhyEvaler, which uses GPT-4o to score generated images based on both scene accuracy and physical correctness.


\noindent
\textbf{PhyGenBench.}
PhyGenBench~\cite{meng2024towards} is a benchmark designed to evaluate the physical commonsense capabilities of text-to-video generation models. It includes 160 carefully crafted prompts that reflect 27 physical laws across the same domains as in PhyBench~\cite{meng2024phybench}, i.e., mechanics, optics, thermal, and material properties.
To support automatic assessment, the authors propose PhyGenEval, a hierarchical evaluation framework that combines vision-language models and GPT-4o for scoring in terms of semantic alignment and physical commensense alignment. 
PhyGenBench constructs the benchmark for video generation through five steps: (1) {Conceptualization}: Identify key physical commonsense from four types of physical commonsense. (2) {Prompt Engineering}: Craft initial prompts to describe the underlying physical phenomenon. (3) {Prompt Augmentation}: Add additional details to the prompts. (4) {Diversity Enhancement}: Use GPT-4o to perform object substitution on the augmented prompts. (5) {Quality Control}: Review the prompts and associated physical laws for accuracy and relevance.

\noindent
\textbf{VideoPhy.}
VideoPhy\cite{bansal2024videophy} categorizes the physical interaction types into three categories which are {Solid-Solid}, {Solid-Fluid} and {Fluid-Fluid}.  
An example of {Solid-Solid} is the prompt \textit{``Bottle topples off the table,"} where the interaction between \textit{``table"} and \textit{``bottle"} is solid-solid, or \textit{``Water flows down a circular drain"} is the interaction between \textit{``water"} and \textit{``drain"} describing {Solid-Fluid} interaction. Similarly, the prompt \textit{``Rain splashing on a pond"} is a {Fluid-Fluid} interaction. Based on these definitions, VideoPhy constructs a total of 688 captions for the benchmark. 
Through extensive human evaluation across 12 models, including both open (e.g., CogVideoX, OpenSora) and closed (e.g., Pika, Gen-2) models, the authors find that most text-to-video models fail to generate physically coherent outputs.
To enable scalable evaluation, the authors propose VideoCon-Physics, a fine-tuned video-language model judging physical plausibility and semantic alignment. 

\noindent
\textbf{VideoPhy2.} 
VideoPhy2~\cite{VideoPhy2} advances the scope of physical commonsense evaluation in text-to-video generation compared to VideoPhy. 
VideoPhy2 shifts toward action-centric evaluation, covering 197 diverse real-world actions, including sports, physical activities, and object manipulations, across 3940 prompts. 
The new dataset introduces longer, more descriptive prompts and augments them with fine-grained annotations of specific physical rules and laws, such as gravity, conservation of mass, and momentum. VideoPhy2 employs a 5-point Likert scale for human evaluations.  
It also offers an enhanced automatic evaluator, VideoPhy2-AutoEval, which demonstrates alignment with human judgments.

\noindent
\textbf{WISA-32K.} WISA-32K~\cite{wang2025wisa} is a curated video dataset designed to enhance the physical understanding of text-to-video generation models. 
It contains 32,000 video clips that depict 17 distinct physical phenomena across three core physics domains, i.e., dynamics, thermodynamics, and optics.  
Each video is annotated with structured physical information, including a textual physical description explaining the underlying principles, qualitative physics categories identifying the types of phenomena present, and quantitative properties such as density, temperature, and time ranges. 
These annotations are automatically derived using GPT-4o mini and Qwen2-VL. The dataset is constructed through video collection, scene detection, aesthetic filtering, and structured annotation.
Experimental results demonstrate that models trained with WISA-32K outperform those trained on general datasets in producing videos that align more closely with real-world physical laws.

\subsubsection{Visual-Conditioned Benchmarks}

Different from text-conditioned benchmarks, visual-conditioned benchmarks involve the evaluation of the predictions of next frames based on visual comparision of physical properties. 


\noindent
\textbf{Physics-IQ.}
Physics-IQ~\cite{PhysicsIQ} is a visual-conditioned benchmark explicitly designed to evaluate the extent to which video generative models understand physical principles through video prediction. Unlike benchmarks that rely on textual descriptions or synthetic data, Physics-IQ leverages a real-world, high-resolution dataset composed of 396 videos filmed under controlled conditions. 
Each video depicts physical interactions governed by principles from solid mechanics, fluid dynamics, thermodynamics, optics, and magnetism. The evaluation task is visual-based, where the models are given an initial conditioning segment and predict the next 5 seconds of the sequence. 
The outputs are then directly compared to ground-truth future frames using spatial, temporal, and pixel-based metrics.
These metrics assess not only where and when motion occurs, but also how much motion is present and how accurately it matches the physical trajectory in the real scene.

\noindent
\textbf{PisaBench.} 
PisaBench~\cite{li2025pisa} is a benchmark introduced to evaluate the physical accuracy of video generative models through the specific task of modeling object freefall.
The benchmark challenges models to generate videos that realistically depict an object falling and potentially colliding with other objects, given only a single input image of the object in midair. 
The authors curate a dataset composed of 361 real-world slow-motion videos and a set of simulated videos generated using the Kubric engine.
PisaBench includes three core evaluations, i.e., using trajectory L2, Chamfer Distance, and Intersection over Union (IoU), which respectively measure the accuracy of object motion, shape fidelity, and spatial consistency over time.

\noindent
\textbf{PhyCoBench.} 
PhyCoBench~\cite{chen2025physical} is a benchmark evaluating the physical coherence of text-to-video generation models. 
The benchmark includes 120 carefully crafted prompts that reflect seven fundamental categories of physical phenomena, including gravity, collision, vibration, friction, fluid dynamics, projectile motion, and rotation. 
These prompts are inspired by physics textbooks and action recognition datasets, with comprehensive coverage of observable motion scenarios.
The authors evaluated four state-of-the-art models through manual rankings based on how realistically the generated videos captured physical behavior.
PhyCoBench also introduces PhyCoPredictor, an automated optical flow-guided frame prediction model that uses the first video frame and the associated prompt to forecast motions and video frames. 
By comparing predictions with the outputs from text-to-video models, PhyCoPredictor assesses physical plausibility using a score based on optical flow and frame reconstruction errors.


\subsubsection{Physical Commonsense Ontology}

One of the main challenges in the evaluation is the inconsistency in defining physical commonsense. Different lines of works have different ways of definition resulting in difficulties for benchmarking between models. This is one of the key questions that the community should focus on in this field. 

COSMOS-Reason1 \cite{Cosmos-Reason1} provides an ontology dividing the physical commonsense into three categories, which are space, time and fundamental physics, with sub-categories in Table \ref{tab:cosmosbench}.
To evaluate the understanding about physical commonsense, COSMOS-Reason1 \cite{Cosmos-Reason1} collected 5737 questions which have 2828 binary questions and 2909 multiple-choice questions. 
Among these questions, a subset of 604 questions are associated with 426 video are manually selected to represent physical commonsense benchmark in which 336 are binary questions and 268 are multiple choice questions.


\begin{table}[htpb!]
  \centering
  \begin{tabular}{|l|r|}
  \hline
  \textbf{Category} & \textbf{Sub-Category} \\
  \hline
  Space & Relationship, Plausibility, Affordance, Environment \\
  \hline
  Time & Actions, Order, Causality, Camera, Planning \\
  \hline
  Fundamental Physics & Attributes, States, Object Permanence, Mechanics, \\
                      & Electromagnetism, Thermodynamics, Anti-Physics \\
  \hline
  \end{tabular}  
  \caption{Taxonomy of physical commonsense in COSMOS-Reason1 \cite{Cosmos-Reason1}.}
  \label{tab:cosmosbench}
\end{table}

\subsection{Evaluation metrics}

There are three main approaches for the evaluation metrics of physical awareness.
The first approach is based on human evaluation, where humans will give the score to the image or video in terms of physical commonsense with defined criteria. 
The second way is based on the perception of VLMs. 
The third way is to use an automatic quantitative score.

\noindent \textbf{Human Evaluation.}
Human-based evaluation is the most direct and reliable method for assessing the physical plausibility of generated visual content. 
In this setup, human annotators are asked to judge whether a generated image or video adheres to the laws of physics, matches a given prompt, or displays coherent dynamics. 
Evaluations can be conducted via rating scales (e.g., binary, three points, or five points) or rankings across different models. 
For instance, in PhyCoBench~\cite{chen2025physical}, humans rank outputs from text-to-video models for each prompt according to the physical realism. 
PhyGenBench~\cite{meng2024towards} uses manual evaluation that includes semantic alignment and physical commonsense alignment, while VideoPhy~\cite{bansal2024videophy} and VideoPhy2~\cite{VideoPhy2} introduce human annotations on semantic adherence, physical commonsense, and the grounding of physical rules.
Though costly and time-consuming, human evaluation remains essential for establishing reliable ground truth.

\noindent \textbf{VLM-Based Evaluation.}
Evaluations based on vision-language models automate human judgment by leveraging large multimodal models to assess physical plausibility through language-guided reasoning. 
These models are either prompted or finetuned to analyze visual content in context of textual descriptions or physics-related questions. 
For example, PhyBench~\cite{meng2024phybench} uses GPT-4o in its PhyEvaler framework to provide item-specific judgments about scene correctness and physical commonsense, mimicking human assessment. 
PhyGenBench~\cite{meng2024towards} takes this further with PhyGenEval, a multi-stage framework where VLMs detect key physical events, verify temporal order, and assess realism through guided questions and frame comparisons. 
VideoPhy~\cite{bansal2024videophy} and VideoPhy2~\cite{VideoPhy2} finetune vision language models using human annotations to predict semantic adherence score, physical commonsense score, and physical rule classification.
These VLM-driven approaches offer scalable and cost-efficient alternatives to manual scoring, though their performance still depends on the quality and specificity of prompts or fine-tuning.
Moreover, the physical understanding ability of VLMs themselves is still an open question.

\noindent \textbf{Automatic Quantitative Evaluation.}
This category involves computational metrics that analyze visual content, focusing on low-level or structured measurements of motion, shape, or physical consistency. 
These metrics are particularly useful for measuring the coherence and predictability of dynamic processes in video prediction. 
For instance, PhyCoBench~\cite{chen2025physical} introduces PhyCoPredictor, which uses optical flow predictions to detect physical anomalies by comparing predicted and generated motion fields. 
PISA~\cite{li2025pisa} employs trajectory tracking (L2 distance), shape matching (Chamfer Distance), and object permanence (IoU) to evaluate whether falling object simulations align with expectations. 
Physics-IQ~\cite{PhysicsIQ} defines a suite of physical understanding metrics, including Spatial and Spatiotemporal IoU, Weighted Spatial IoU, and Mean Squared Error (MSE), which quantify where, when, and how much motion occurs in generated videos compared to real-world references. These metrics are precise, reproducible, and ideal for benchmarking, though they usually require paried real-world videos as the ground truth.

\section{Discussions}

In this section, we discuss related concepts and provide comparisons with physics-aware generation.

\noindent
\textbf{Semantic Awareness vs. Physical Awareness.}
Semantic awareness and physical awareness serve distinct but complementary roles in how models interpret and generate visual data.
Semantic awareness involves interpreting and recognizing concepts and objects within visual data, focusing on identifying ``what" and ``where" are present in a scene, by learning \textit{static} knowledge mapping from pixels to latent features.
In contrast, physical awareness pertains to the ability to simulate \textit{dynamic} interactions and transitions governed by physical laws, such as motion, collisions, forces, and material behaviors, addressing the ``how" and ``why" of visual data. 
In this sense, physics understanding is characterized by its dynamic and predictive nature, which explains why temporal modeling of videos is especially important in this domain.
Physics-aware generation aims to combine both aspects to produce content that is both semantically meaningful and physically accurate.

\noindent
\textbf{Geometry Awareness vs. Physical Awareness.}
Geometry awareness involves perceiving and modeling the \textit{extrinsic} structural properties of objects and scenes. This includes aspects such as shape, size, pose, depth, and positions\cite{Bahmani2024a,Yang2024d,Chou2024,Bahmani2024b,Zhu2024c}.
Physical awareness focuses on modeling the \textit{intrinsic} properties and behaviors of objects and environments. This involves reasoning about how objects move, deform, and interact under physical laws\cite{Li2023b}.
In this sense, physical awareness is characterized by its causality and interaction modeling.
Integrating both geometry and physics is essential for creating embodied models that not only perceive but also understand and interact with the physical world.


\section{Future Directions}

In this section, we discuss several potential future directions along the promise of physics-aware generation. 


\noindent
\textbf{Better Evaluation of Physical Awareness.}
Although standardized benchmarks tailored to physical awareness have been introduced recently\cite{Meng2024}, the evaluation remains ad-hoc or largely relies on manual assessment.
The evaluation of physical consistency and dynamic realism in generative models remains a significant challenge that hinders progress.
One potential direction is to integrate generative models with established physics engines to evaluate physical plausibility, by converting and deploying generated content within simulation platforms.
Incorporating task-oriented evaluations can also be beneficial to assessing physical reasoning capabilities. For example, assessing how well generated content performs in downstream tasks, such as object manipulation in robotics or environmental modeling in autonomous driving, can offer feedback on a model's utility and physical validity.
It is also promising to explore the alignment between generated content and the physical world using embodied agents as an evaluation measurement, for example, by using the behavior similarity of embodied agents running in the simulated and the real world.

\noindent
\textbf{Explainability by Physical Awareness.}
The integration of physics-aware mechanisms into models holds immense potential for enhancing model explainability.
Incorporating explicit physical laws into generative models can offer interpretable pathways linking inputs to outputs, enabling users to trace how specific forces, constraints, and interactions govern the evolution of generated content. 
By mapping model decisions to well-understood physical principles, the generation process becomes more transparent and aligned with human intuition.
Combining physics-aware generative models with existing interpretability methods, such as saliency maps\cite{GradCAM} and feature attribution\cite{kim2018interpretability}, can further enhance explainability. For example, visualizing how physical parameters influence specific regions of generated outputs can reveal the model's reliance on physical properties during generation.

\noindent
\textbf{Physics-Augmented Foundational Models.}
It is promising to augment large foundational models with physical understanding and utilize them for physics-aware generation\cite{Gao2024c}\cite{Balazadeh2024}.
This includes a versatile understanding of physical world by unifying various perception foundation models\cite{Liu2023a,Yu2023a,Ren2024a,Yang2024a,Ravi2024,Khirodkar2024,Yang2023a,Wang2024f}, and enhanced reasoning abilities from large language models augmented with 3D\cite{Hong2023,Zhen2024} and physical knowledge\cite{Cherian2024}.
Another future direction is physics-guided pretraining and representation learning, by leveraging large-scale synthetic datasets and differentiable physics integration.
It is also interesting to explore self-supervised physical learning, by implementing learning objectives that encourage the discovery of physical laws in large models.

\noindent
\textbf{Neural-Symbolic Hybrid Models.}
The integration of neural and symbolic reasoning presents a promising direction for advancing physics-aware generation, where symbolic systems can offer rigorous reasoning based on structured knowledge of physics and causality\cite{Hsu2023a,Hsu2023b}.
Embedding symbolic physical constraints, such as differentiable temporal logic\cite{xu2022don}, can guide generative models to produce outputs that are physically valid and temporally consistent.
Symbolic representations also excel in compositional reasoning, allowing systems to understand complex entities and relationships. 
By incorporating symbolic graphs or ontologies, generative models can achieve more complex compositional generation.

\noindent
\textbf{Generative Simulation Engine.}
Generative simulation engine offers the potential to generate physical simulation directly from textual or semantic prompts, with Genesis\cite{Genesis} as a potential example. 
Such text-to-simulation models aim to convert high-level textual descriptions into fully interactive, physics-consistent virtual environments. 
For instance, a command such as ``simulate a landslide on a steep mountain after heavy rain" could trigger the generation of a realistic 3D terrain, modeled with appropriate soil and fluid dynamics, and simulate the resulting landslide using physics-based solvers.
Generative simulation engines can automate and speed up the design of physical simulation, which can be further integrated into larger systems for physics-aware generation in various modalities.
The current main obstacle for generative simulation is its limited generalizability across diverse text prompts and scenes.

\noindent
\textbf{Robotics and Embodied AI.}
A natural application of physics-aware generation is to be used as world models or simulators to train robots and embodied agents.
By generating physically realistic synthetic data that accurately reflects real-world dynamics, robots can be trained more effectively in simulation and seamlessly be transited to real-world operations\cite{Xue2024a,Sun2024a,Akkerman2024}.
It is also interesting to explicitly inject physical reasoning capabilities into Vision-Language-Action models\cite{Cheang2024,Zhen2024} to better predict actions and outcomes in complex and transferred environments with physical knowledge that can better generalize.

\noindent
\textbf{Applications in Interdisciplinary Domains.}
Physics-aware generation also holds transformative potential across many application domains.
It can improve climate modeling by enhancing the fidelity, resolution, and scale of simulations\cite{Earth2}.
In healthcare, physics-aware generation can significantly enhance surgical training and planning, such as building surgical simulations that accurately replicate the physical properties of tissues to provide realistic training environments\cite{Yang2024c,Wang2024b}.

\section{Conclusion}

This survey comprehensively reviewed the rapidly evolving field of physics-aware generation in computer vision, highlighting the efforts made to enhance the physical realism and functionality of generated visual content.
The core of this survey focused on the integration of physical simulation into generative models, delineating between physics-aware generation with explicit simulation and without explicit simulation.
Physics-aware generation currently stands at a transformative juncture, bridging the gap between virtual and physical realities. 
We hope that this survey serves as a helpful resource and inspiration for future research in the field.

\ifCLASSOPTIONcaptionsoff
  \newpage
\fi



%



\bibliographystyle{IEEEtran}
\bibliography{paper}

\end{document}